\def\year{2020}\relax
\documentclass[letterpaper]{article} %
\usepackage{aaai20}  %
\usepackage{times}  %
\usepackage{helvet} %
\usepackage{courier}  %
\usepackage[hyphens]{url}  %
\usepackage{graphicx} %
\usepackage{graphicx}  %
\nocopyright
\newcommand{\citet}[1]{\citeauthor{#1}~\shortcite{#1}}
\newcommand{\citep}{\cite}

\usepackage{amsmath,amssymb}
\usepackage{adjustbox}
\usepackage{algorithm}
\usepackage{algorithmic}
\usepackage{amsmath, amssymb, mathtools}
\usepackage{bm}
\usepackage{booktabs}
\usepackage{cleveref}
\usepackage{comment}
\usepackage{todonotes}

\crefname{equation}{eq.}{eqs.}

\DeclareMathOperator{\MI}{I}
\DeclareMathOperator{\Entr}{H}
\DeclareMathOperator{\KL}{D_{KL}}
\DeclareMathOperator{\Expect}{\mathbb{E}}
\DeclareMathOperator{\Embedding}{\mathbf{e}}

\title{Improved Variational Neural Machine Translation \\ by Promoting Mutual Information}
\author{Arya D. McCarthy\textsuperscript{\rm 1,2} \and Xian Li\textsuperscript{\rm 2} \and Jiatao Gu\textsuperscript{\rm 2} \and Ning Dong\textsuperscript{\rm 2} \\
    \textsuperscript{1}Johns Hopkins University\\
    \textsuperscript{2}Facebook\\
    \texttt{arya@jhu.edu}, \{\texttt{xianl},\texttt{jgu},\texttt{dnn}\}\texttt{@fb.com}
}

\begin{document}

\maketitle

\begin{abstract}

\begin{comment}
Variational neural machine translation (NMT) (and more generally, variational modeling for text) is largely an unsolved problem. This paper proposes a simple and effective approach of improving the variational model performance, especially in dealing with the ``mode collapse'' problem that is commonly observed in VAE for text.
We present a new latent variable model for neural machine translation, extending both Transformer and conditional variational autoencoders. We successfully prevent the degeneration problem of \textbf{posterior collapse}, by (1) proposing and optimizing a modified evidence lower bound (ELBO) objective which promotes mutual information between the latent variable and the target, and (2) informing the information content of the latent variable with an auxiliary bag-of-words prediction task. Empirical results show that our model effectively utilizes the latent variable, avoiding posterior collapse. Further, the results improve translation quality and robustness when translating noisy datasets. We also demonstrate the proposed model is capable of utilizing source-side monolingual data to boost translation performance.

indicating that variational models could be one effective way of dealing with MT training data noise

We alleviate the posterior collapse problem while improving translation quality.

We translate between English and three languages: Romanian, Sinhala, and French.
\end{comment}
Posterior collapse plagues VAEs for text, especially for conditional text generation with strong autoregressive decoders. In this work, we address this problem in variational neural machine translation by explicitly promoting mutual information between the latent variables and the data. Our model extends the conditional variational autoencoder (CVAE) with two new ingredients: first, we propose a modified evidence lower bound (ELBO) objective which explicitly promotes mutual information; second, we regularize the probabilities of the decoder by mixing an auxiliary factorized distribution which is directly predicted by the latent variables. We present empirical results on the Transformer architecture and show the proposed model effectively addressed posterior collapse: latent variables are no longer ignored in the presence of powerful decoder. As a result, the proposed model yields improved translation quality while demonstrating superior performance in terms of data efficiency and robustness.
\end{abstract}

\section{Introduction}

Neural sequence-to-sequence models with attention have become the \textit{de facto} methods for machine translation~\cite{bahdanau2014neural,vaswani2017attention}. NMT models require a large amount of parallel data to surpass the quality of phrase-based statistical models, and they are very sensitive to data quality~\cite{koehn2017six}. As a conditional text generation task, machine translation contains both \textit{intrinsic} uncertainty, where a given sentence usually has multiple valid reference translations, and \textit{extrinsic} uncertainty, due to noise in the sentence alignment that produces parallel training data~\cite{ott2018analyzing}. %

As an option for handling data uncertainty, latent variable models such as variational autoencoders (VAE) have been investigated in language modeling and conditional text generation~\cite{miao2016neural,zhang2016variational,yang2017improved}. However, in contrast to their success when applied to computer vision tasks~\cite{kingma2013auto,rezende2014stochastic}, VAEs in natural language processing suffer from \textit{posterior collapse}, where the learnt latent code is ignored by the decoder~\cite{bowman2015generating}. 

In this work, we propose to address posterior collapse when using latent variable models in neural machine translation. First, we provide an analysis of the  evidence lower bound (ELBO) used in conditional variational autoencoders (CVAE) commonly used in conditional text generation. Our analysis reveals that optimizing CVAE's ELBO not only inevitably leads to vanishing divergence of the posterior from the prior during training, but also decreasing mutual information between latent codes and data. Based on this insight, we propose two modifications of CVAE's ELBO to address this problem: 1) we explicitly add mutual information back to the training objective in a principled way, and 2) we use a factorized decoder, predicting ``bag of words" as an auxiliary decoding distribution to regularize latent variables, finding that both are complementary. We summarize our contribution as follows:
\begin{enumerate}
    \item %
    We improve CVAE by enhancing mutual information between latent variables and data, effectively mitigating posterior collapse in conditional text generation.
    \item %
    We apply the proposed model in neural machine translation with the  Transformer architecture. Experiments demonstrate that latent variables are not ignored even in the presence of the powerful autoregressive decoder. Compared to variational NMT with CVAE architecture or non-latent Transformer, the proposed improvements yield improved robustness and data-efficiency.
    \item We extend the proposed model to semi-supervised learning with monolingual data, and show that it has superior performance on self-training by effectively learn from source-side monolingual data.
\end{enumerate}
\begin{figure*}
\begin{center} 
\includegraphics[width=\textwidth]{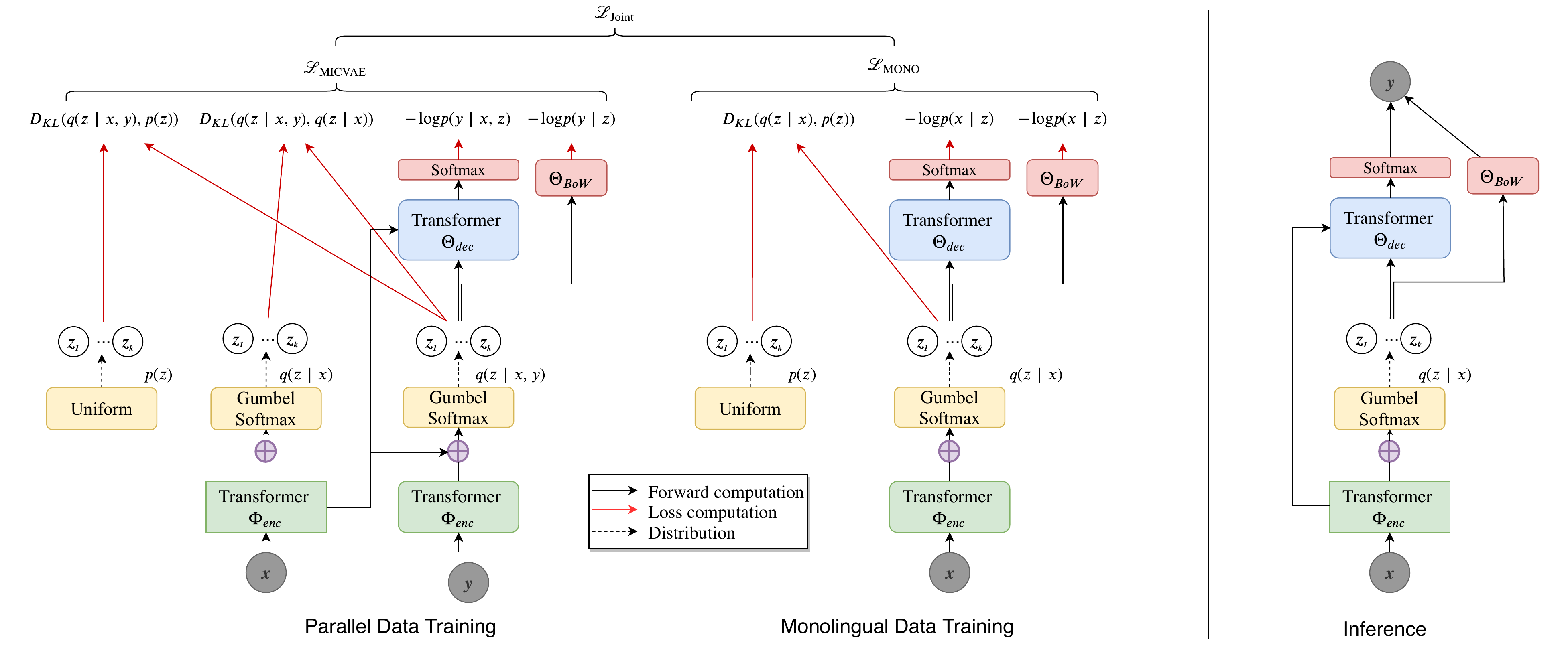}
\end{center}
\caption{Model architecture, including training with only parallel data, and joint training with monolingual data.}
\label{fig:training}
\end{figure*}
\section{Background}

\subsection{Neural Machine Translation} 
Problem instances in machine translation are pairs of sequences \((\bm{x} \triangleq [x_1, \ldots, x_m], \bm{y} \triangleq [y_1, \ldots, y_n])\), where \(\bm{x}\) and \(\bm{y}\) represent the source and target sentences, respectively.
Conventionally, a neural machine translation (NMT) model is a parameterized conditional distribution whose likelihood factors in an autoregressive fashion:
\begin{equation}
    p_\theta\left(\bm{y}\mid\bm{x}\right) = \prod_{i}^{|\bm{y}|} p_\theta\left(y_i \mid \bm{x}, \bm{y}_{<i}\right)\text{.}
\end{equation}
The dominant translation paradigm first represents the source sentence as a sequence of contextualized vectors (using the \emph{encoder}), then decodes this representation token-by-token into a target hypothesis according to the above factorization. 
The parameters \(\theta\) are learned by optimizing the log-likelihood of training pairs with stochastic gradient methods \citep{bottou2004large}. Decoding the model occurs in a deterministic fashion, using an efficient approximate search like beam search \citep{tillmann-ney-2003-word}. Recently, Transformer with multi-head attention has become the state of the art for NMT \citep{vaswani2017attention}.

\subsection{Conditional Variational Autoencoder (CVAE)}
Our NMT approach extends the conditional variational autoencoder (CVAE) \citep{sohn2015learning}, of which variational NMT \citep{zhang2016variational} is a particular case. It introduces a latent variable $\bm{z}$ to model the conditional distribution:
\begin{equation}
    p_\theta(\bm{y} \mid \bm{x}) = \int_{\bm{z}}p_\theta(\bm{y}\mid \bm{z}, \bm{x}) \cdot p(\bm{z} \mid \bm{x})\, \mathrm{d}z
    \text{.}
    \label{eqn:log-likelihood}
\end{equation}
However, it is intractable to directly marginalize \(\bm{z}\). Instead, the CVAE objective is to maximize the \textbf{evidence lower bound (ELBO)} of the \mbox{(log-)}likelihood:
\begin{multline}
    \mathcal{L}_{\mathrm{CVAE}}(\phi, \theta; \bm{x}, \bm{y}) = \Expect_{q_{\phi}(\bm{z}\mid \bm{x}, \bm{y})} \left[\log p_\theta(\bm{y}\mid \bm{x}, \bm{z})\right] \\
    - \KL(q_{\phi}(\bm{z}\mid \bm{x}, \bm{y}) \parallel p_\theta(\bm{z} \mid \bm{x}))
    \text{,}
    \label{eqn:cvae}
\end{multline}
where $\KL$ represents the Kullback--Leibler (KL) divergence between two distributions.
Learning is done by amortized variational inference, where the variational distribution \(q_{\phi}(\bm{z}\mid \bm{x}, \bm{y})\) is an inference network parameterized by \(\phi\). 

\subsection{Posterior Collapse}
Posterior collapse can be explained mathematically by analysis of the ELBO objective, as well as from the perspective of a powerful decoder. We consider both in this subsection.

We first provide an analysis of CVAE's objective and identify its drawback. Recall that our computed loss approximates the loss on the true data distribution by using a finite number of samples:
\begin{equation}
    \mathcal{L} = \Expect_{p_{\mathcal{D}}(\bm{x}, \bm{y})} \left[ \mathcal{L}_{\mathrm{CVAE}}(\phi, \theta; \bm{x}, \bm{y}) \right]
\end{equation}
Thus, the KL term is:
\begin{align}
    &\Expect_{p_\mathcal{D}(\bm{x}, \bm{y})} \left[\KL(q_\phi(\bm{z} \mid \bm{x}, \bm{y}) \parallel p_\theta(\bm{z} \mid \bm{x})) \right] \nonumber \\
    &\triangleq  \Expect_{p_\mathcal{D}(\bm{x}, \bm{y})}\Expect_{q_{\phi}(\bm{z}\mid \bm{x}, \bm{y})} \left[\log q_{\phi}(\bm{z}\mid \bm{x}, \bm{y}) - \log p(\bm{z}\mid \bm{x})\right] \nonumber \\
    &=     \sum_{\bm{x}, \bm{y}} q(\bm{x}, \bm{y}, \bm{z}) \log \frac{q (\bm{x}, \bm{y}, \bm{z})}{p(\bm{x}, \bm{y}, \bm{z})} \nonumber \\
    &= \Expect_{\bm{x}, \bm{y}, \bm{z}}\log \frac{q(\bm{x}, \bm{y} \mid \bm{z}) q(\bm{z})}{p(\bm{x}, \bm{y}) p(\bm{z})} \nonumber \\
    &= \Expect_{q_\phi(\bm{z})}\Expect_{q_\phi(\bm{x}, \bm{y} \mid \bm{z})} \log \frac{q(\bm{x}, \bm{y} \mid \bm{z})}{p_\mathcal{D}(\bm{x}, \bm{y})} + \Expect_{q_\phi(\bm{x}, \bm{y}, \bm{z})} \log \frac{q(\bm{z})}{p(\bm{z})} \nonumber \\
    &= \underbrace{-\Entr(\bm{x}, \bm{y} \mid \bm{z}) + \Entr(\bm{x}, \bm{y})}_{\triangleq \MI_{q_{\phi}}(\bm{z}; \bm{x}, \bm{y})} %
    + \underbrace{
        \Expect_{q_\phi(\bm{z})} \log \frac{q(\bm{z})}{p(\bm{z})}
        }_{\triangleq \KL(q_{\phi}(\bm{z}) \parallel p(\bm{z} ) )}  \label{eqn:kl}
\end{align}

The third line comes from multiplying the numerator and denominator by \(p_\mathcal{D}(\bm{x}, \bm{y})\) following \citet{hoffman2016elbo}, the fact that \(p(\bm{z} \mid \bm{x})\) is conditionally independent of \(\bm{y}\), and defining \(p_{\mathcal{D}}(\bm{x}, \bm{y}) \triangleq \frac{1}{N}\) for all \(N\) training samples \((\bm{x}, \bm{y}) \in \mathcal{D}\). The fifth line comes from factoring and conditional independence.

As the two resulting terms are non-negative \citep{cover-thomas-2006-elements}, the global minimum of  \Cref{eqn:kl} is \(\MI_{q_{\phi}}(\bm{z}; \bm{x}, \bm{y}) = \KL(q_{\phi}(\bm{z}) \parallel p(\bm{z} ) ) = 0 \). Unfortunately, at this point, the consequence of the optimization is that \(\bm{z}\) is conditionally independent of the data.

Another explanation of posterior collapse is the ``powerful decoder" perspective: an autoregressive model with large capacity comes to approximate a complex distribution \emph{without using the latent variables} \cite{bowman2015generating,he2019lagging}. This is a challenge for NMT, which requires a powerful decoder such as Transformer with direct attention to the encoder. %

\section{Addressing Posterior Collapse}

\subsection{CVAE Guided by Mutual Information}

\subsubsection{Adding $\MI_{q_{\phi}}(\bm{z}; \bm{x},\bm{y})$ to ELBO}

To combat the optimization dilemma from \cref{eqn:kl}, we explicitly add the mutual information term to the CVAE's ELBO and obtain a new training objective:
\begin{multline}
\label{eq:micvae}
    \mathcal{L}_{\mathrm{MICVAE}} =\mathcal{L}_{\mathrm{CVAE}} + \MI_{q_{\phi}}(\bm{z}; \bm{x}, \bm{y}) \\ = 
     \Expect_{q_{\phi}(\bm{z}\mid \bm{x}, \bm{y})}\log p(\bm{y}\mid \bm{x}, \bm{z}) - \KL(q_{\phi}(\bm{z}) \parallel p(\bm{z} ) )
    \text{.}
\end{multline}
The new training objective aims to match the aggregated posterior distribution of the latent variable $q_{\phi}(\bm{z})$ to the aggregated prior distribution $p(\bm{z})$. It can be seen as an extension of InfoVAE~\citep{zhao2017infovae} to conditional generative models, where we have overcome the mismatch between the (joint) data distribution \(p_\mathcal{D}(\bm{x}, \bm{y})\) and the (conditional) log-likelihood objective \(p_\theta(\bm{y} \mid \bm{x})\). %

\subsubsection{Guiding $\bm{z}$ to Encode Global Information}

Several existing approaches weaken the decoder to encourage latent variables to be utilized, which is not preferred in practice \citep{bowman2015generating,guljarani2016pixelvae}. Here we propose a different approach: explicitly guiding the information encoded in $\bm{z}$ without reducing the decoder's capacity.  

Inspired by an information-theoretic view of posterior collapse using Bits-Back Coding theory~\cite{wallace-freeman-1987-estimation,Hinton:1993:KNN:168304.168306,chen2016variational}, we add an auxiliary loss for $\bm{z}$ to encode information which cannot be modelled locally by the autoregressive decoder distribution $\prod_t p_\theta(y_t \mid \bm{x}, \bm{y}_{<t})$. We use bag-of-words (BoW) prediction as the auxiliary loss. %
It encodes global information while having a non-autoregressive factorization $\prod_t p_\psi(y_t \mid \bm{z})$. The auxiliary decoder complements the autoregressive decoder (which is locally factorized) by combining predictions at the Softmax layer, i.e.\ $p(y_t \mid \bm{x}, \bm{y}_{<t}, \bm{z})$ is a \textbf{mixture of softmaxes} \citep{yang2018breaking}:
\begin{multline}
        p(y_t \mid \cdot) = (1-\lambda) \cdot p_{\theta}(y_t \mid \bm{x}, \bm{y}_{<t},  \bm{z}) \\ 
        + \lambda \cdot p_{\psi}(y_t \mid \bm{z})
    \text{.}
\end{multline}
Thus, the bag-of-words objective regularizes the log-likelihood bound.

\subsection{Architecture}
\paragraph{ Inference Network} We use discrete latent variables with reparameterization via Gumbel-Softmax~\cite{jang2016categorical} to allow backpropagation through discrete sampling. Compared to the multivariate Gaussian distribution commonly used in VAE and CVAE, our parameterization allows us to explicitly account for multiple modes in the data. To make our model more general, we introduce a \emph{set} of discrete latent variables \(\bm{z} = \{\bm{z}_1, \ldots, \bm{z}_K\}\) which are independently sampled from their own inference networks $\Phi_k$. Specifically, each $\Phi_k$ computes dot product attention with encoder outputs  $\bm{h}\in \mathbb{R}^d $:
\begin{equation}
\bm{C}_k = \text{Softmax}(\frac{\bm{e}_{k}\bm{W}^k(\bm{h}\bm{W}^h)^\top}{\sqrt{d}})\bm{h}\bm{W}^h
\text{.}
\end{equation}
We can now sample $\bm{z}_k$ by Gumbel-Softmax reparameterization trick~\cite{jang2016categorical}:
\begin{equation}
\begin{split}
\bm{z}_k = \text{GumbelSoftmax}(\textbf{C}_k) 
=\text{softmax}\left(\frac{\textbf{C}_k + \bm{g}}{\tau}\right),
\end{split}
\end{equation}
where $\bm{g}=-\log(-\log(\bm{u})), \bm{u}\sim \text{Uniform}$ is the Gumbel noise and $\tau$ is a fixed temperature (we use $\tau=1$ in this paper). In the inference time, we use a discrete version by directly sampling from the latent variable distribution.

\paragraph{BoW Auxiliary Decoder}
Given inferred sample  $\bf{z} \sim \Phi_k({\textbf{h}})$, the BoW decoder predicts all tokens at once without considering their order.  
We compute the cross-entropy loss for the predicted tokens over the output vocabulary space \(V\): 
\begin{equation}
\mathcal{L}_{\mathrm{BoW}} = \sum_{i=i}^{|V|} p_i \log \hat{p_\psi}(y_i \mid \bm{z}), \quad \sum_{i=1}^{| V| } p_i = 1
\text{.}
\end{equation}
We take the empirical distribution $p_i$ to be a token's frequency within a sentence normalized by its total frequency within a mini-batch, mitigating the effect of frequent (stop) words. $\hat{p}_{
\psi}%
$ is computed by conditioning on the latent code only, without direct attention to encoder outputs. We use dot-product attention between the latent embeddings and the token embeddings (each of dimensionality \(d\)):
\begin{equation}
\label{eq:bow_loss}
p_{\psi}(y_i \mid \bm{z}) = \text{Softmax}_i \left(\frac{\Embedding(\bm{z})\Embedding^T(V)}{\sqrt{d}}\right)
\text{.}
\end{equation}

\subsection{Training}
\label{sec:model_training}
We train our model using amortized variational inference, where samples $\bm{z}$ are drawn from the posterior distributions to get a Monte Carlo estimate of the gradient. In addition to standard CVAE supervised learning with parallel data, we also extend our model to be jointly trained by adding monolingual data.

\todo[disable,author={R1}]{Instead of adding the extra objective Iq(z;x) from the source monolingual data, why not also consider Iq(z;y) from the target monolingual data? Since the posterior distribution q(z|x,y) is conditioned on both source and target, it would be better to promote mutual info in both directions.}

\todo[disable,author={R1}]{How did you compute the mutual info I(z;x)? Better say one or two sentences about the evaluation method.
}

\paragraph{Semi-supervised learning} We apply the same modification to VAE's ELBO, following \citet{zhao2017infovae}. For jointly training with source-side monolingual data, we add $\MI_{q_{\phi}}(\bm{z}; \bm{x})$ to the ELBO\footnote{Learning to copy the source text has proven useful for low-resource NMT \citep{currey-etal-2017-copied}.}, and for target-side monolingual data, we add  $\MI_{q_{\phi}}(\bm{z}; \bm{y})$. %
The joint objective sums the modified CVAE and VAE objectives%
:
\begin{equation}
\label{eq:mono_loss}
\begin{split}
   \mathcal{L}_{\mathrm{Mono}} = & \log p(\bm{x} \mid \bm{z}) \\
    &+{}  D_{\mathrm{KL}}\left(\frac{1}{N} \sum_{n=1}^N  q_{\phi}(z_n | x_n) \;\bigg{|\bigg|}\; \frac{1}{N} \sum_{n=1}^N  p(z_n)\right) \\
\end{split}
\end{equation}
\begin{equation}
    \label{eq:joint_loss}
    \mathcal{L}_{\mathrm{Joint}} =  \mathcal{L}_{\mathrm{MICVAE}} + \mathcal{L}_{\mathrm{Mono}}
\end{equation}
\Cref{alg:main} describes the overall training strategy. 
\begin{algorithm}[t]
\caption{\label{alg:main}Training Strategy}
\begin{algorithmic}[1]
    \STATE $\Phi_{enc}, \Phi_{k=1, ..., K}, \Theta_{dec},\Theta_{BoW} \gets \text{initialize parameters}$ 
    \WHILE{$\Theta_{enc}, \Theta_{dec},\Theta_{BoW}, \Phi_{k=1, ..., K}$ have not converged}
		\STATE{Sample $(\mathbf{x}, \mathbf{y})$ from $D^{\text{bitext}} $} 
		\STATE{Compute $\mathcal{L}_{\mathrm{MICVAE}}$ with \Cref{eq:micvae}}
		\STATE{Train $\Phi_{enc}, \Theta_{dec}, \Phi_{k=1, ..., K}$ with $\mathcal{L}_{\mathrm{MICVAE}}$}
		\STATE{Compute $\mathcal{L}_{\mathrm{BoW}}$ with \Cref{eq:bow_loss}}
		\STATE{Train $\Phi_{enc},\Theta_{BoW}, \Phi_{k=1, ..., K}$ with $\mathcal{L}_{\mathrm{BoW}}$}
		\IF{\text{self\_training}}
		\STATE{Sample $\mathbf{x}$ from $D^{\text{mono}} $} 
		\STATE{Compute $\mathcal{L}_{\mathrm{Mono}}$ with \Cref{eq:mono_loss}}
		\STATE{Train $\Phi_{enc}, \Phi_{k=1, ..., K}$ with $\mathcal{L}_{\mathrm{Mono}}$}
		\ENDIF
	\ENDWHILE
\end{algorithmic}
\end{algorithm}

\section{Experiments}

Here we describe our experiments, showing that our techniques have practical value for both mitigating posterior collapse and improving translation quality.
\subsection{Setup}
\paragraph{Datasets}
First, we evaluate our models on standard WMT benchmark datasets. Second, we focus on two representative challenges in NMT: low-resource and robustness to noisy data.
\begin{description}
    \item[WMT14 German--English] We use data from the WMT14 news translation shared task, which has 3.9M sentence pairs for training with BPE tokenization.   
    \item[WMT16 Romanian--English] We use data from the WMT16 news translation shared task. We use the same BPE-preprocessed \citep{sennrich-etal-2016-neural} train, dev and test splits as in \citet{gu2017non} with 608k sentence pairs for training. 
    \item[Low resource benchmark (FLoRes) Sinhala--English] We use the same preprocessed data as in \citet{guzman2019two}. There are 646k sentence pairs. 
    \item[MT for Noisy Text (MTNT) French--English] We use 30K subword units built jointly from source and target sentences, and only keep sentences with less than 100 tokens. For training, there are 34,380 sentence pairs for English-French and 17,616 sentence pairs for French--English \cite{michel2018mtnt}. We also used 18,676 \emph{monolingual} sentences per language from the same data source (Reddit).
\end{description}
\paragraph{Implementation} All of our models are implemented using Transformer architecture.%
For WMT14 De--En and WMT16 Ro--En%
, we use the base configuration \citep{vaswani2017attention}: 6 blocks, with 512-dimensional embedding, 2048-dimensional FFN, and 8 attention heads. For FLoRes (low-resource) and MTNT (both low-resource and noisy), we use a smaller Transformer: 4 layers, 256-dimensional embedding, 1024-dimensional inner layers, and 4 attention heads. %
Input and output embeddings are shared between the inference network and decoder. We use $T=4$ categorical latent variables each of dimension 16 which are found by grid search over validation set.  Auxiliary bag-of-words predictions are combined with decoder prediction with  $\lambda=0.1$.  All models are optimized using Adam with $\beta_1=0.9$, $\beta_2=0.98$, $\epsilon=1e-8$, weight decay of 0.001, and the same warmup and learning rate schedule as in \citet{ott2018scaling}. All models are trained on 8 \textsc{Nvidia} V100 GPUs with 32K tokens per mini-batch. We train WMT14 De-En with 200k updates and all other models with 100k updates.

We employ joint BPE vocabularies. The sizes are 32k for En--De and En--Ro; 30k for Fr--En; and 3k for Si--En. In addition, we use a word dropout rate of 0.4 during training of the baseline and latent variable models, which is complementary to our approach.

\paragraph{Baselines} We compare our model to three baselines: 1) \textit{Transformer, non-latent}: standard Transformer model without latent variables (denoted as non-latent), 2) \textit{VNMT}: CVAE model with Gaussian distribution as was proposed in Variational NMT by \citet{zhang2016variational}, which we reimplemented using Transformer, and 3) \textit{DCVAE}: CVAE model with the same discrete latent variables parameterization as ours but without the proposed enhancement on promoting mutual information, i.e., the only differences are the modified ELBO and bag-of-words regularizer.  

\section{Main Results}
\subsection{Preventing Posterior Collapse}

\begin{figure*}[t]
\begin{center}
\includegraphics[width=\textwidth]{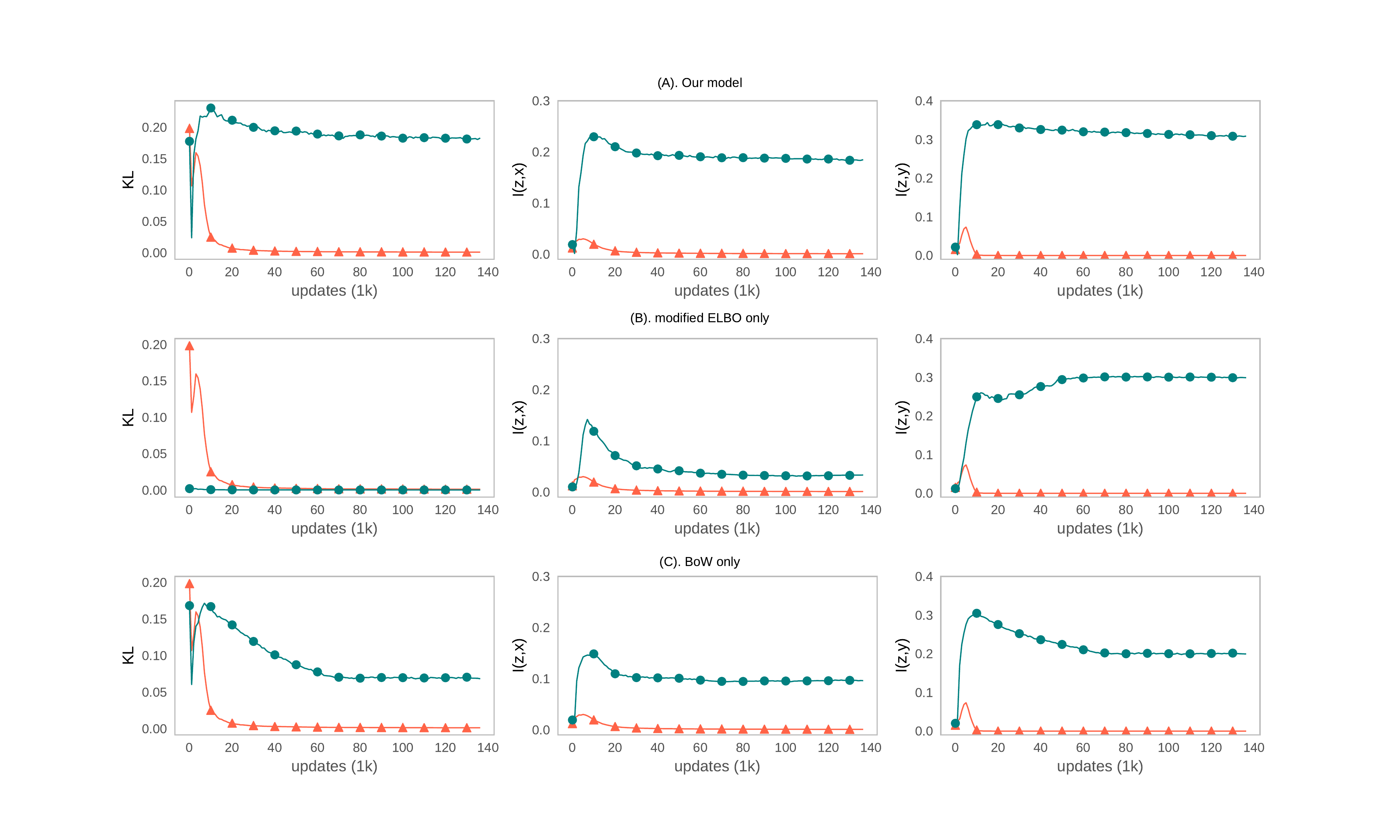}
\end{center}
\caption{Row (A): comparison of KL and mutual information between baseline (DCVAE, solid triangle, orange color) and our model (solid circle, teal color). Row (B) and (C): ablation study on relative contribution from MICVAE and BoW. All metrics are computed on WMT16 Ro--En validation set during training.}
\label{fig:kl_mi}
\end{figure*}

In this set of experiments, we compare our model to a standard DCVAE without the proposed enhancement in mutual information.  We report four metrics of posterior collapse on validate set of WMT Ro--En:
\begin{enumerate}
     \item Kullback--Leibler divergence (KL).
    \item Mutual information between the latent variable and the data: $\MI_{q_{\phi}}(\bm{z}, \bm{x})$ and $\MI_{q_{\phi}}(\bm{z},\bm{y})$.
    \item Negative log-likelihood (NLL) per token.
\end{enumerate}
   
\Cref{tab:collapse_metrics} shows that when using standard DCVAE ELBO, even with the common practice of KL annealing (KLA), both the KL loss and mutual information settle to almost 0 which is consistent with the analysis in \Cref{eqn:kl}. We also plot the progression of \(\KL\), \(\MI_{q_{\phi}}(\bm{z}; \bm{x})\), and \(\MI_{q_{\phi}}(\bm{z}; \bm{y})\) during training in \Cref{fig:kl_mi}. The posterior collapse of the baseline model is apparent: both \(\KL\) mutual information terms drop to 0 at the beginning of training as a result ELBO's design. On the other hand, our model, without using any annealing schedule, can effectively increase mutual information and prevent KL loss from settling to a degenerate solution early on.

\todo[disable,author=R3]{The comparison of KL term in table 1 seems meaningless, because the term in different methods are different.}

\todo[disable,author={Arya}]{Which language is \Cref{tab:collapse_metrics} describing?}
\todo[disable,author={Xian}]{Added in the table caption}
\begin{table}
\caption{Results on improving posterior collapse. The KL value refers to $\KL(q_{\phi}(\bm{z}\mid \bm{x}, \bm{y}) \parallel p(\bm{z} \mid \bm{x} ))$ for DCVAE and $\KL(q_{\phi}(\bm{z}\mid \bm{y} ) \parallel p(\bm{z} \mid \bm{x} ))$ for our model.}
\smallskip
\centering
\begin{adjustbox}{max width=\linewidth}
\begin{tabular}{l r r r r}
\toprule
Model & \(\KL\)  & $\MI_{q_{\phi}}(\bm{z},\bm{x})$ &   $\MI_{q_{\phi}}(\bm{z},\bm{y})$ & NLL  \\
\midrule
DCVAE + KLA & 0.001 & 0.001 & 4.2\textsc{e}-6  & 3.17  \\
Our model & 0.17 & 0.18 & 0.31 & 3.16  \\

\bottomrule
\end{tabular}
\end{adjustbox}
\label{tab:collapse_metrics}
\end{table}

\begin{comment}
\begin{figure*}
\begin{center}
\includegraphics[width=0.9\textwidth]{kl-ablation.pdf}
\end{center}
\caption{Ablation study on relative effects of modifed ELBO only, and BoW only on KL and mutual information on WMT16 Ro--En validation set during training.}
\label{fig:kl-ablation}
\end{figure*}
\begin{figure*}
\begin{center}
\includegraphics[width=0.9\textwidth]{kl-mi-bow_only.pdf}
\end{center}
\caption{Effect of BoW only on KL and mutual information on WMT16 Ro--En validation set during training.}
\label{fig:bow-only}
\end{figure*}
\end{comment}
\todo[disable]{In \Cref{tab:collapse_metrics}, it seems the proposed methods makes NLL worse? Also, just by using the modified ELBO, the KL becomes even smaller than CVAE baseline?}
\todo[disable,author={R2}]{As mutual information is a measure of the amount of information contained in one random variable about another random variable, how can it be the KL decrease while the I increase(since claimed the property of KL in VAE setting, the smaller, the better)?}
\todo[disable,author={R3}]{Table 1 and figure 2 show that even with Modified ELBO and BOW, the KL terms are still at a very small value. In my experiments wrt VNMT personally, such low KL terms seemed to contribute very little information to the decoders. \textbf{I suggest authors to provide more analysis of whether the z thing does actually play an important role in your model}, such as substituting z with a zero vector.}
\todo[disable,author={R3}]{It seems that the bag-of-words decoder plays a much more important role than modified ELBO according to the results in table 1. I think you need to evaluate the contribution of the modified ELBO to your model. Does modified ELBO improves BLEU score ?}

\subsection{Translation Quality}

We report corpus-level BLEU \cite{papineni2002bleu}\footnote{Particularly, we use detokenized SacreBLEU \citep{post-2018-call}.} on the test sets where the translations are generated by sampling each $z_k$ with soft-assignment (vs. argmax). %
\todo[disable]{Are the BLEU scores reported the mean score of several sampled translations (from the latent space)? }

\todo[disable, author={R2}]{Sincerely, it's good to report real result, but maybe you should explain why on the WMT16 EN-RO dataset, your model performs slightly worse, as VNMT is supposed to learning better due to the newly added information.}

\paragraph{Supervised Learning on Parallel Data} First, we evaluate our model's performance when trained with parallel data on standard WMT datasets. \Cref{parallel_results} shows that our model consistently outperforms both VNMT  and DCVAE models---which requires ad-hoc KL annealing (KLA) while on par with  a strong Transformer baseline. %

\begin{table}
\caption{BLEU score on WMT benchmarks.}
\smallskip
\centering
\adjustbox{max width=\linewidth}{
\begin{tabular}{l r r r r }
\toprule
 & \multicolumn{2}{c}{WMT16} & \multicolumn{2}{c}{WMT14} \\  
 \cmidrule(lr){2-3} \cmidrule(lr){4-5} %
Model & Ro--En & En--Ro & De--En & En--De \\ %
\midrule
VNMT & 34.20 & 34.27 & 30.35 & 25.84 \\
DCVAE & 34.16 & 34.51 & 29.76 & 25.46 \\
Our model & 34.76 & 34.97 & 31.39 & 26.42  \\
\midrule
Non-latent & 34.73 & 34.54 & 30.89 & 26.36  \\
  \bottomrule
\end{tabular}}
\label{parallel_results}
\end{table}
\begin{comment}

\paragraph{Low resource NMT: fully supervised}
Next, we evaluate our model on low-resource scenarios which is an unsolved challenge in NMT \citep{koehn2017six}. \Cref{lowres} summarizes the results on two representative low-resource datasets.
\begin{table}
\caption{BLEU score on low-resource datasets.}
\smallskip
\centering
\adjustbox{max width=\linewidth}{
\begin{tabular}{l r r r r }
\toprule
 & \multicolumn{2}{c}{MTNT} & \multicolumn{2}{c}{FLORES} \\  
 %
 \cmidrule(lr){2-3} \cmidrule(lr){4-5} \\ %
Model & Fr--En & En--Fr & Si--En & En--Si \\ %
\midrule
%
%
Non-latent & 26.65 & running & TODO & TODO  \\
%
VNMT & 26.90 & running & TODO & TODO \\
DCVAE & 26.37 & running & TODO & TODO \\
%
Our model & \textbf{28.58} & running & TODO & TODO  \\
  \bottomrule
\end{tabular}}
\label{lowres}
\end{table}

\end{comment}
\paragraph{Semi-supervised with Source-side Monolingual Data}
Leveraging monolingual data is a common practice to improve low resource NMT. Current approach has been mostly focusing on using target-side monolingual data through ``backtranslation" as a data augmentation, while how to effectively leverage source-side monolingual to facilitate self training is still an open challenge \citep{sennrich2015improving,zhang2016exploiting}. We use the joint training objective described in \Cref{eq:joint_loss}. To have a fair comparison, we also extend VNMT and DCVAE with the same joint training algorithm, i.e., the newly added monolingual data is used to train their corresponding sequence encoder and inference network with standard VAE ELBO. That is, the only difference is that our model was trained to promote mutual information $\MI_{q_{\phi}}(\bm{z}, \bm{x})$ and $\MI_{q_{\phi}}(\bm{z}, \bm{y})$. As shown in \Cref{table:mono}, by doing so the proposed model brings larger gains during self-training with source-side monolingual data.

\begin{table}
\caption{Translation performance (BLEU) of utilizing source-side monolingual data.}
\smallskip
\centering
\begin{adjustbox}{max width=\linewidth}
\begin{tabular}{l r r}
\toprule
Model & Fr--En & En--Fr  \\
\midrule
DCVAE & 26.37 & 26.11 \\
+ source mono & 27.30 &  26.40  \\
Our model & 28.58 & 26.31 \\
+ source mono & 29.81 & 26.69 \\
  \bottomrule
\end{tabular}
\end{adjustbox}
\label{table:mono}
\end{table}

\begin{figure}
    \centering
    \includegraphics[width=\linewidth]{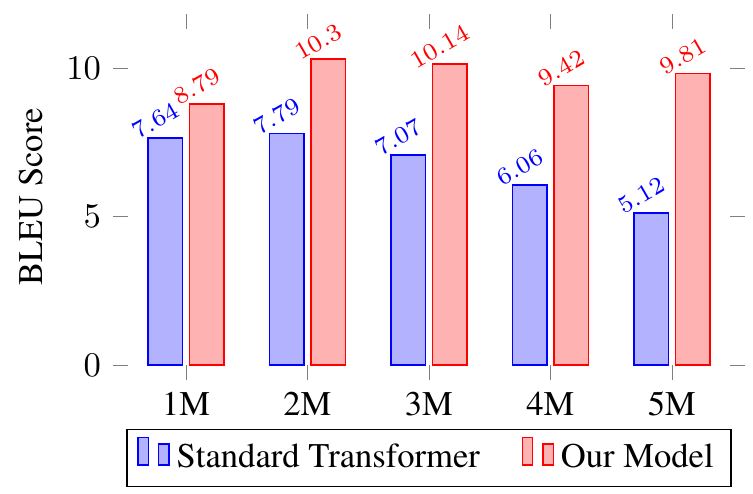}
    \caption{BLEU when increasing the amount of noisy parallel data in training, Si--En.}
    \label{fig:si_en}
\end{figure}

\paragraph{Robustness to noisy data}
While high-quality parallel data is scarce for low-resource language pairs, weakly aligned sentence pairs can be mined from massive unpaired data such as Paracrawl\footnote{\url{https://paracrawl.eu/}}. We evaluate our model's performance when augmenting the training set with increasingly noisy parallel data filtered by Zipporah \cite{xu2017zipporah}. \Cref{fig:si_en} %
shows the results in the Sinhala--English direction. Our model always outperforms standard Transformer, which struggles as more (and noisier) data is added.

\todo[disable, author={R2}]{on the comparison between with and without mono corpus, they compare to a NON-latent model, which is not fair, as least they should list their results on a unmodified VAE model, i.e., this experiment result is not convincible.}

\todo[disable, author={R3}]{Experiment about monolingual data and noisy data lacks comparison with VNMT.}

\todo[disable, author={R2}]{I doubt that their model performs better on the noise data just like what word dropout does(which has been proven to be effective).
}

\begin{comment}

\paragraph{Data Efficiency}
We evaluate the proposed model's sample-efficiency by varying the amount of training data. As is shown in \Cref{table:data-effiency}, \todo[inline]{plot this result.} 

\begin{table}
\caption{Translation performance (BLEU) with increasing amount of parallel data.}
\smallskip
\centering
\begin{adjustbox}{max width=\linewidth}
\begin{tabular}{l r r r}
\toprule
Model & De-En 500k & De-En 1M & De-En 3.9M  \\
\midrule
Non-latent & 20.89 & 24.08 &30.89 \\
DCVAE & 20.54 & 23.70 & 29.76 \\
Our model & 21.20 & 24.26 & 31.39 \\
  \bottomrule
\end{tabular}
\end{adjustbox}
\label{table:data-effiency}
\end{table}
\end{comment}
\section{Analysis}
\subsection{Ablation Study}
We further investigate how different ingredients of our proposed approach contribute to preventing posterior collapse and improving translation quality. We conduct further experiments with two variants of the proposed model: 1) modified ELBO only: only adding mutual information term to the training objective, while without gradients from $\mathcal{L}_{\mathrm{BoW}}$, 2) BoW only: which is equivalent to DCVAE combined with Bow decoder. 

First, we perform the same collapse metrics evaluation as in \Cref{tab:collapse_metrics}. \Cref{fig:kl_mi} (B) suggests that by explicitly adding mutual information term back to the training objective, both  $\MI_{q_{\phi}}(\bm{z}, \bm{x})$ and $\MI_{q_{\phi}}(\bm{z}, \bm{y})$ are effectively raised, while the remaining aggregated KL term is still optimized to zero. Such behavior is consistent with the analysis revealed in \Cref{eqn:kl}. On the other hand, regularizing $z$ with BoW decoder only, as is shown in \Cref{fig:kl_mi} (C), is very effective in preventing KL vanishing as well as increasing mutual information. When two approaches are combined, as was shown in \Cref{fig:kl_mi}, the model retain higher mutual information for both  $\MI_{q_{\phi}}(\bm{z}, \bm{x})$ and $\MI_{q_{\phi}}(\bm{z}, \bm{y})$.

Next, we look into whether such difference in mutual information lead to difference in translation quality. We compare these two models: BoW only (\Cref{fig:kl_mi} (C)) and both (\Cref{fig:kl_mi} (A)) on WMT14 De--En and WMT16 Ro--En test sets. \Cref{table:ablation-bleu} reveals that such difference matters more in low-data regime.
\begin{table}
\caption{Ablation study on translation quality (BLEU).}
\smallskip
\centering
\begin{adjustbox}{max width=\linewidth}
\begin{tabular}{l r r}
\toprule
Model & De--En (3.9M)  & Ro--En (608K)   \\
\midrule
Both & 31.39 & 34.76  \\
BoW only &31.14 & 34.22 \\
  \bottomrule
\end{tabular}
\end{adjustbox}
\label{table:ablation-bleu}
\end{table}

\subsection{Analysis of Outputs}

Delving into model predictions helps us understand how our model outperforms the others. We provide some 1-best predictions from the Romanian--English data in \Cref{tab:outputs}.
Several examples support the fact that our model has more fluent and accurate translations than the baseline or VNMT. VNMT often struggles by introducing disfluent words, and both VNMT and the baseline can select justifiable but incorrect words. For instance, in our second example, the gender and animacy of the possessor are not specified in Romanian. Our model selects a more plausible pronoun for this context.

More broadly, we find that the reference translations are quite loose and context-dependent (rather than word-for-word translations), making it difficult for models to reproduce---they give reasonable translations with greater fidelity to source word order and content. (As an extreme example, the English translation of \emph{ed miliband isi cunostea dusmanii} adds information to the beginning: \emph{for all his foolishness ed miliband knew who his enemies were}; no model is able to add this.) Our model often makes superior judgments in terms of lexical choice and fluency.

\begin{table}[t]
    \centering
    \caption{Translation examples from the baseline Transformer, VNMT, and our model. Disfluent words or absences are marked in \textcolor{red}{red}, and slightly incorrect lexical choice is marked in \textcolor{blue}{blue}. Romanian diacritics have been stripped.}
    \smallskip
    \begin{adjustbox}{max width=\linewidth}
\begin{tabular}{l} 
    \toprule
	\textbf{Source}: ma intristeaza foarte tare .\\
	\textbf{Reference}: that really saddens me . \\
	\textbf{Base}: i am very saddened .\\
	\textbf{VNMT}: i am saddened very \textcolor{red}{loudly} . \hfill\emph{(Wrong sense of \emph{tare})}\\
	\textbf{Ours}: i am very saddened .\\
	\midrule
	\textbf{Source}: cred ca executia sa este gresita .\\
	\textbf{Reference}: i believe his execution is wrong .\\
	\textbf{Base}: i believe that \textcolor{blue}{its} execution is wrong .\\
	\textbf{VNMT}: i believe that \textcolor{blue}{its} execution is wrong .\\
	\textbf{Ours}: i believe that his execution is wrong .\\
	\midrule
	\textbf{Source}: da , chinatown\\
	\textbf{Reference}: yes , chinatown\\
	\textbf{Base}: yes , chinatown\\
	\textbf{VNMT}: yes , \textcolor{red}{thin} \textcolor{blue}{.}\\
	\textbf{Ours}: yes , chinatown\\
	\midrule
	\textbf{Source}: nu stiu cine va fi propus pentru aceasta functie .\\
	\textbf{Reference}: i do not know who will be proposed for this position .\\
	\textbf{Base}: i do not know who will be proposed for this \textcolor{blue}{function} .\\
	\textbf{VNMT}: i do not know who will be proposed for this \textcolor{blue}{function} .\\
	\textbf{Ours}: i do not know who will be proposed for this position .\\
	\midrule
	\textbf{Source}: recrutarea , o prioritate tot mai mare pentru companii\\
	\textbf{Reference}: recruitment , a growing priority for companies\\
	\textbf{Base}: recruitment , \textcolor{blue}{an increasing} priority for companies\\
	\textbf{VNMT}: recruitment , \textcolor{red}{[article missing]} increasing priority for companies\\
	\textbf{Ours}: recruitment , a growing priority for companies\\
	\bottomrule
			\end{tabular}
			\end{adjustbox}
    \label{tab:outputs}
\end{table}
\subsection{Analysis of Latent Variables}
\begin{figure}[t]
\begin{center}
\includegraphics[width=\linewidth]{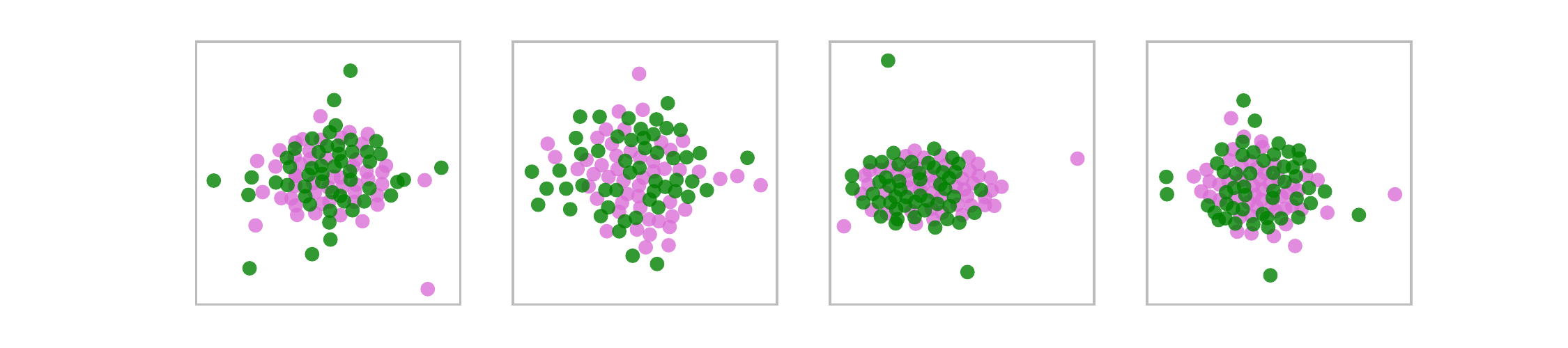}
\end{center}
\caption{t-SNE visualization of $\bm{z}_k$, $k=1,2,3,4$ samples inferred from 100 sentences from two datasets with distinct domains, MTNT (orchid) and WMT news (green).}
\label{fig:z_tsne}
\end{figure}

Finally, we probe whether different latent variables encode different information. We random sample 100 sentences from two test sets of distinct domains, MTNT (Reddit comments) and WMT (news) with 50 sentences each. We plot the t-SNE projection of their corresponding latent variables samples $z_k$ inferred from $\Phi_k$, $k=1,2,3,4$ respectively. Figure \ref{fig:z_tsne} indicates that different latent variables learn to organize the data in different manners, although there was no clear signal that any of them exclusively specialize in encoding a domain label. We leave an thorough analysis of their information specialization to future work. 

\section{Related Work}
Unlike most prior work in (conditional) text generation, we are able to address posterior collapse without requiring an annealing schedule \citep{bowman2015generating}, a weakened decoder \citep{guljarani2016pixelvae}, or a restriction on the variational family \citep{razavi2018preventing}. 

Unlike \citet{ma-etal-2018-bag}, who also employ bag-of-words as an objective for NMT, our bag-of-words decoder only has access to \(\bm{z}\), not the encoder states. Conversely, unlike \citet{weng-etal-2017-neural}, our generative decoder has access to both the latent variable and the encoder states, and the bag-of-words prediction is handled by a separate set of parameters.

Posterior collapse for text VAE was first identified in language modeling \cite{bowman2015generating}. %
VNMT \cite{zhang2016variational} applies CVAE with Gaussian priors to conditional text generation. VRNMT \cite{su2018variational} extends VNMT by modeling the translation process in greater granularity. All of them needed manually designed annealing schedules to increase KL loss to mitigate posterior collapse. Discrete latent variables have been applied to NMT \cite{gu2017non,shen2019mixture,kaiser2017one} but did not use variational inference or address posterior collapse. Tackling posterior collapse has received more attention lately, with general approaches such as aggressively trained inference networks \cite{he2019lagging}, skip connections \cite{dieng2018avoiding}, and more expressive priors \cite{razavi2018preventing,tomczak2017vae}.

\section{Conclusion}
We have presented a conditional generative model with latent variables whose distribution is learned with variation inference, then applied it to the task of machine translation. Our approach does not require an annealing schedule or a hamstrung decoder to avoid posterior collapse. Instead, by providing a new analysis of the conditional VAE objective to improve it in a principled way and incorporating an auxiliary decoding objective, we measurably rely on the latent variables. In addition to preventing posterior collapse, our approach improves translation quality in terms of BLEU. Empirical evaluation demonstrates that the proposed method has improved performance in dealing with uncertainty in data, including weakly supervised learning from source-side monolingual data as well as noisy parallel data.

\bibliography{li}
\bibliographystyle{aaai}
\end{document}